\pdfoutput=1
\documentclass[letterpaper, 10 pt, conference]{ieeeconf}

\IEEEoverridecommandlockouts

\usepackage{graphics} %
\usepackage{epsfig} %
\usepackage{mathptmx} %
\usepackage{times} %
\usepackage{amsmath} %
\usepackage{amssymb}  %
\usepackage{cite}
\usepackage{gensymb}
\usepackage{multirow}
\usepackage{booktabs}
\usepackage{graphicx} %
\usepackage{subcaption} %
\usepackage{color}
\usepackage{float}
\usepackage{arydshln}
\usepackage{float}
\usepackage{hyperref}

\usepackage{xargs}
\usepackage{xcolor}  %
\usepackage[colorinlistoftodos,prependcaption,textsize=tiny]{todonotes}
\newcommandx{\Sen}[2][1=]{\todo[inline,linecolor=red,backgroundcolor=red!25,bordercolor=red,#1]{#2}}

\title{\LARGE \bf
        Digital Beamforming Enhanced Radar Odometry
}

\author{Jingqi Jiang$^{1}$, 
        Shida Xu$^{1,2}$, 
        Kaicheng Zhang$^{1,2}$,
        Jiyuan Wei$^{1}$,
        Jingyang Wang$^{3}$ and 
        Sen Wang$^{1}$%
        \thanks{$^{1}$I-X and Department of Electrical and Electronic Engineering,
        Imperial College London, UK
                        {\tt\small
                        \{j.jiang23, s.xu23, k.zhang23, j.wei23,
                        sen.wang\}@imperial.ac.uk}}%
        \thanks{$^{2}$School of Engineering and Physical Sciences, Heriot-Watt University,
        UK {\tt\small \{sx2000,kz13\}@hw.ac.uk}}%
        \thanks{$^{3}$Department of Electronic Engineering, Tsinghua University,
        China {\tt\small wangjingyang.2019@tsinghua.org.cn}}%
}

\begin{document}

\maketitle
\thispagestyle{empty}
\pagestyle{empty}

\begin{abstract}

        Radar has become an essential sensor for autonomous navigation,
        especially in challenging environments where camera and LiDAR sensors
        fail.
        4D single-chip millimeter-wave radar systems, in particular, have drawn
        increasing attention thanks to their ability to provide
        spatial and Doppler information with low hardware cost and power consumption.
        However, most single-chip radar systems using traditional signal processing, such as Fast Fourier Transform, suffer from
        limited spatial resolution in radar detection, %
        significantly limiting the performance of radar-based odometry and
        Simultaneous Localization and Mapping (SLAM) systems.
        In this paper, we develop a novel radar signal processing
        pipeline that integrates spatial domain beamforming techniques,
        and extend it to 3D
        Direction of Arrival estimation.
        Experiments using public datasets are conducted to evaluate and
        compare the performance of our proposed signal processing
        pipeline against traditional methodologies.
        These tests specifically focus on assessing structural
        precision across diverse scenes and measuring odometry
        accuracy in different radar odometry systems.
        This research demonstrates the feasibility of achieving more accurate radar odometry by simply replacing the standard FFT-based processing with the proposed pipeline. %
        \renewcommand{\thefootnote}{\fnsymbol{footnote}}
        The codes are available at \href{https://github.com/SenseRoboticsLab/DBE-Radar}{GitHub}\footnote{https://github.com/SenseRoboticsLab/DBE-Radar}.

\end{abstract}

\section{INTRODUCTION}
Autonomous robots often perceive their environments using a
combination of sensors, such as LiDAR\cite{zhangCURLMAPContinuousMapping2024}
and cameras\cite{jiangDVIOOptimizationBasedTightly2021}, for autonomous navigation.
However, these optical sensors tend to be limited by environmental factors,
such as adverse weather.
In contrast, {radio detection and ranging} (radar) technology
is known for its efficacy
in various environmental conditions, being an appealing sensor
for autonomous navigation systems.

Among radar systems,
Frequency Modulated Continuous Wave (FMCW) millimeter-wave radar systems
have gained popularity in autonomous driving applications
due to their ability to simultaneously provide distance and, importantly, velocity measurements
\cite{venonMillimeterWaveFMCW2022,
        harlowNewWaveRobotics2024}.
However, %
the excess noise and the low resolution in the Direction of Arrival (DoA)
measurements greatly affect the quality of radar detection,
degrading the performance of radar-based odometry and SLAM systems.
Therefore, in early radar SLAM systems, they primarily used 2D mechanical
scanning radar sensors
\cite{hongRadarSLAMRobustSimultaneous2022,
        adolfssonLidarlevelLocalizationRadar2023,
        parkPhaRaODirectRadar2020,
        mockRadaRaysRealtimeSimulation2023}.
These radar sensors employ mechanical spinning
mechanisms, similar to LiDAR sensors,
to concentrate power into a narrower beam for
increased radar gain and reduced sidelobes, enhancing DoA estimation.
However, they are bulky and unable to handle three-dimensional
space effectively, which restricts their wide adoption of  %
ground-based autonomous vehicles.

Recently, 4D System-on-Chip (SoC) radar systems have emerged
as a promising solution for robotic navigation applications
\cite{han4DMillimeterWaveRadar2024}.
Without the need for mechanical spinning, these radar systems
leverage Multi-Input-Multi-Output (MIMO) technology to estimate a target's
DoA
and Doppler information.
Additionally, they can be miniaturized into a more compact form factor at a fraction of the cost,
facilitating their integration into a wider range of applications.
However,
the compact size of current SoC radar designs inherently restricts the number of physical antennas that can be accommodated,
which reduces the %
accuracy of DoA and then the performance of radar SLAM.

To alleviate the radar SLAM performance degradation, %
some works attempted to use multiple radars \cite{doerXRIORadarInertial2021,huangMultiRadarInertialOdometry2024}
or fuse other sensors
\cite{kubelkaWeNeedScanmatching2024,
        huangLessMorePhysicalEnhanced2024,
        doerEKFBasedApproach2020,
        girodRobustBaroradarinertialOdometry2024}.
However, not only do these approaches fail to
resolve the issue of low DoA accuracy fundamentally,
but they also lead to more complicated
system design and integration
with the introduced additional challenges on sensor
calibration and synchronization.

Two main directions
have been explored in existing literature
to address this issue in radar SLAM.
The first direction focuses on using
multiple sets of antennas from several chips, such as a cascaded 4D SoC radar, to increase aperture.
Consequently, the DoA accuracy and the density
of the radar point cloud can be improved,
opening the door to transfer some LiDAR-based SLAM techniques,
like generalized Iterative Closest Point (ICP), point cloud based loop closure
\cite{zhang4DRadarSLAM4DImaging2023}, and ground
segmentation \cite{chenDRIORobustRadarInertial2023}, to radar SLAM.
However, the cascaded radar solution often results in increased sensor size and power consumption, alongside low frame rate and high computational complexity.

The second direction focuses on enhancing the radar point cloud quality
via postprocessing algorithms.
Some learning-based methods \cite{prabhakaraHighResolutionPoint2023,
        zhangDenseAccurateRadar2024,chengNovelRadarPoint2022}
utilize LiDAR point clouds as supervision
to train models for predicting higher-quality point clouds.
Although these methods could improve the quality of point cloud
of single-chip mmWave radar,
they may not generalize well in environments not being trained
and can not provide the Doppler information of point clouds,
which is required by most radar
odometry/SLAM systems as a critical measurement.
As for the non-learning-based methods, Zeng et al.
\cite{zengAngularSuperResolutionRadar2021}
apply the super-resolution techniques (compressed sensing)
to the radar signal to realize
angular superresolution radar imaging. They show that
super-resolution techniques could improve the localization
result of occupancy grid map in 2D SLAM.
Meiresone et al. \cite{meiresoneEgomotionEstimationLowpower2022}
utilize a capon-based range-azimuth image generation method
before adopting a 2D LiDAR odometry pipeline for radar to estimate the
ego-motion of the vehicle.
However, neither the learning-based nor model-based methods delve into radar signal processing within the context of radar odometry/SLAM.

\begin{figure}[t]
        \centering
        \includegraphics[width=0.47\textwidth]{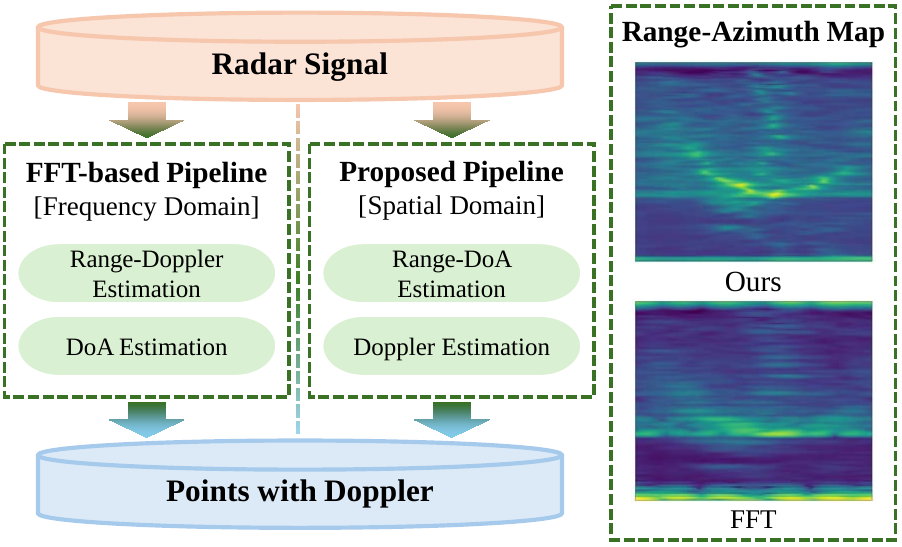}
        \caption{Comparison of radar signal processing pipelines.
        }
        \label{fig:pipepline_comparison}
        \vspace{-0.5cm}
\end{figure}

Inspired by the significance of feature extraction (i.e., extracting information from image data) in visual SLAM,
this work investigates how to leverage raw radar data for radar odometry/SLAM.
Fast Fourier Transform (FFT) method is a widely recognized standard practice for processing raw radar data cube \cite{bilikRiseRadarAutonomous2019},
due to its ability to quickly transform time-domain signals into the frequency domain for
extracting frequency components,
such as Doppler shifts and range information.
A typical full FFT-based radar signal processing pipeline
is illustrated in  Figure \ref{fig:pipepline_comparison}.
First, 2D-FFT is performed on the fast-time and
the slow-time dimensions, generating
range-Doppler maps.
Then the range-Doppler maps of all virtual antennas are
non-coherently accumulated
before applying detection algorithms, like Constant False Alarm Rate (CFAR), to extract potential targets.
Finally, the targets' direction of arrival (DoA)
is estimated by applying FFT on the antenna dimension.
Although the radar point cloud density can be improved by
the non-coherent accumulation, which enhances the signal-to-noise ratio
of weak reflective targets,
its Doppler's estimation bias can be passed to the
DoA estimation, degrading the accuracy of the radar point cloud.
Moreover, the FFT-based pipeline is hampered by sidelobe noise
and low resolution from angle FFT,
which further limits the SLAM performance.
Evolving from the FFT method, this paper proposes a novel digital beamforming-based pipeline to process raw radar data, %
significantly improving the precision of radar points and consequently
the accuracy of radar-based 3D SLAM systems.
Our main contributions are as follows:
\begin{itemize}
        \item A novel radar signal processing pipeline
              that integrates advanced spatial domain beamforming
              techniques and extends them to 3D DoA
              estimation.
        \item Extensive experiments on both filter and graph based radar odometry methods, demonstrating substantial accuracy gain on
              both radar localization and mapping by replacing the FFT pipeline.
\end{itemize}

\begin{figure*}[htpb]
        \centering
        \includegraphics[width=1.\textwidth]{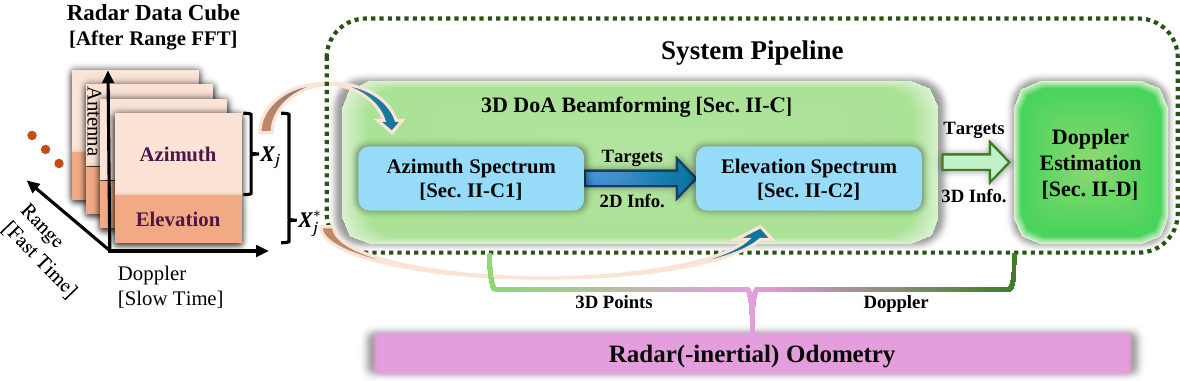}
        \caption{System overview of the proposed pipeline.
        }
        \label{fig:system_overview}
\end{figure*}
\section{METHODOLOGY}

\subsection{System Overview}

The proposed radar signal processing pipeline is
illustrated in Figure \ref{fig:system_overview}.
Different from the frequency domain based traditional pipeline,
our pipeline is based on the spatial domain.
1D FFT is firstly conducted on the fast time dimension
(Section \ref{subsec:mimo}).
Then, the DoA beamforming is performed on the range-FFT results to
estimate the DoA of potential targets in the environment (Section \ref{subsec:doa}).
Given the DoA information, all radar echoes are coherently accumulated
and FFT is performed on the slow time dimension to estimate the Doppler
of the corresponding targets (Section \ref{subsec:Doppler}).

\subsection{MIMO Radar Signal Model}
\label{subsec:mimo}
\begin{figure}[tbp]
        \centering
        \includegraphics[width=0.45\textwidth]{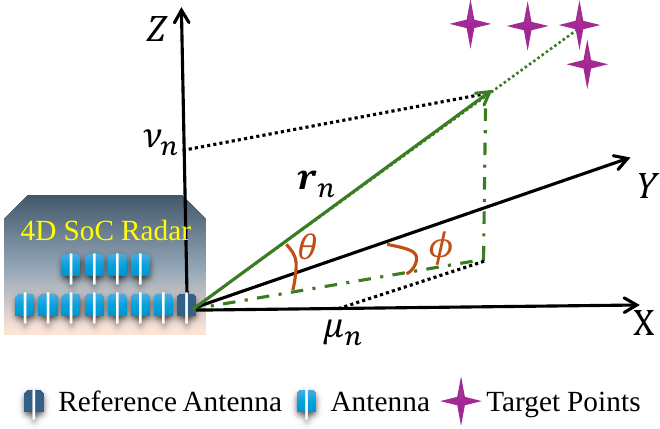}
        \caption{MIMO radar geometric model.}
        \label{fig:mimo_radar_model}
\end{figure}
Based on the concept and principles of 2D MIMO radar in \cite{raoMIMORadar2018},
we derive the 3D MIMO radar signal model in a more general form. As shown in
Figure \ref{fig:mimo_radar_model}, a typical MIMO radar system
consists of multiple virtual antennas along X-axis (azimuth dimension) and
Y-axis (elevation dimension). The potential targets are modeled as
single point scatterers in the environment. These scatterers are located at
the positive Y-axis (range dimension) and have different DoA and Doppler.

The traditional MIMO radar signal model usually uses trigonometric
functions to describe the DoA relationship between
the virtual antennas and the scatterers. However, this representation
makes the formulation of the radar signal model more complex and less
intuitive for programming. In this work, the DoA of the $n$-th scatterer
is represented by a three-dimensional (3D) unit vector as
\begin{equation}
        \label{eq:doa_vector}
        \mathbf{r}_{n} = \begin{bmatrix}
                \mu_{n} & \sqrt{1-\mu^2_{n}-\nu^2_{n}} & \nu_{n}
        \end{bmatrix}^{\top}, \quad  \mu^2_{n} + \nu^2_{n} \leq 1
\end{equation}
where \(\mu_{n}\) and \(\nu_{n}\) are the one-dimensional parameters in
the domain of real numbers.
Note that \(\mu_{n}\) and \(\nu_{n}\) are not the commonly used
azimuth {\(\phi_{n}\)} and elevation \(\theta_{n}\) angles. They can be
converted to the real azimuth and elevation angles by
\begin{equation}
        \label{eq:doa_angle}
        \begin{aligned}
                \theta_{n} & = \sin^{-1}(\nu_{n}),                                 \\
                \phi_{n}  & = \sin^{-1}\left({\mu_{n}}/{\sqrt{1 - \nu^2_{n}}}\right).
        \end{aligned}
\end{equation}
Given the \(m\)-th virtual antenna position
$\mathbf{d}_m$,
the phase difference
between the reference antenna and the \(m\)-th virtual antenna
can be calculated by
\begin{equation}
        \label{eq:phase_difference}
        \begin{aligned}
                {\Delta \varphi_{n,m}} & = 2\pi\mathbf{r}_n^{\top} \mathbf{d}_m / \lambda,     %
        \end{aligned}
\end{equation}
where \(\lambda\) is the wavelength of the radar signal.

Now we define the steering vector of the \(n\)-th scatterer as
\begin{equation}
        \label{eq:steering_vector}
        \mathbf{a}_n = \mathbf{a}(\mu_n,\nu_n)  = \begin{bmatrix}
                e^{j\Delta \varphi_{n,1}} & e^{j\Delta \varphi_{n,2}} & \ldots & e^{j\Delta \varphi_{n,M}}
        \end{bmatrix}^{\top},
\end{equation}
where \(M\) is the total number of virtual antennas.
{The steering vector \( \mathbf{a}_n \) represents the phase shifts
        of the \(n\)-th scatterer
        observed across the array elements.}
Then, the received signal of all virtual antennas $\mathbf{x}_n$ for
the \(n\)-th scatterer at the \(k\)-th time index
can be represented as
\( \mathbf{x}_n(k) = \mathbf{a}_n{s}_0(k) \), where
\({s}_0(k)\) is the transmitted signal from the target.

\begin{figure*}[thp]
        \centering
        \begin{subfigure}[b]{0.329\textwidth}
                \includegraphics[width=\textwidth]{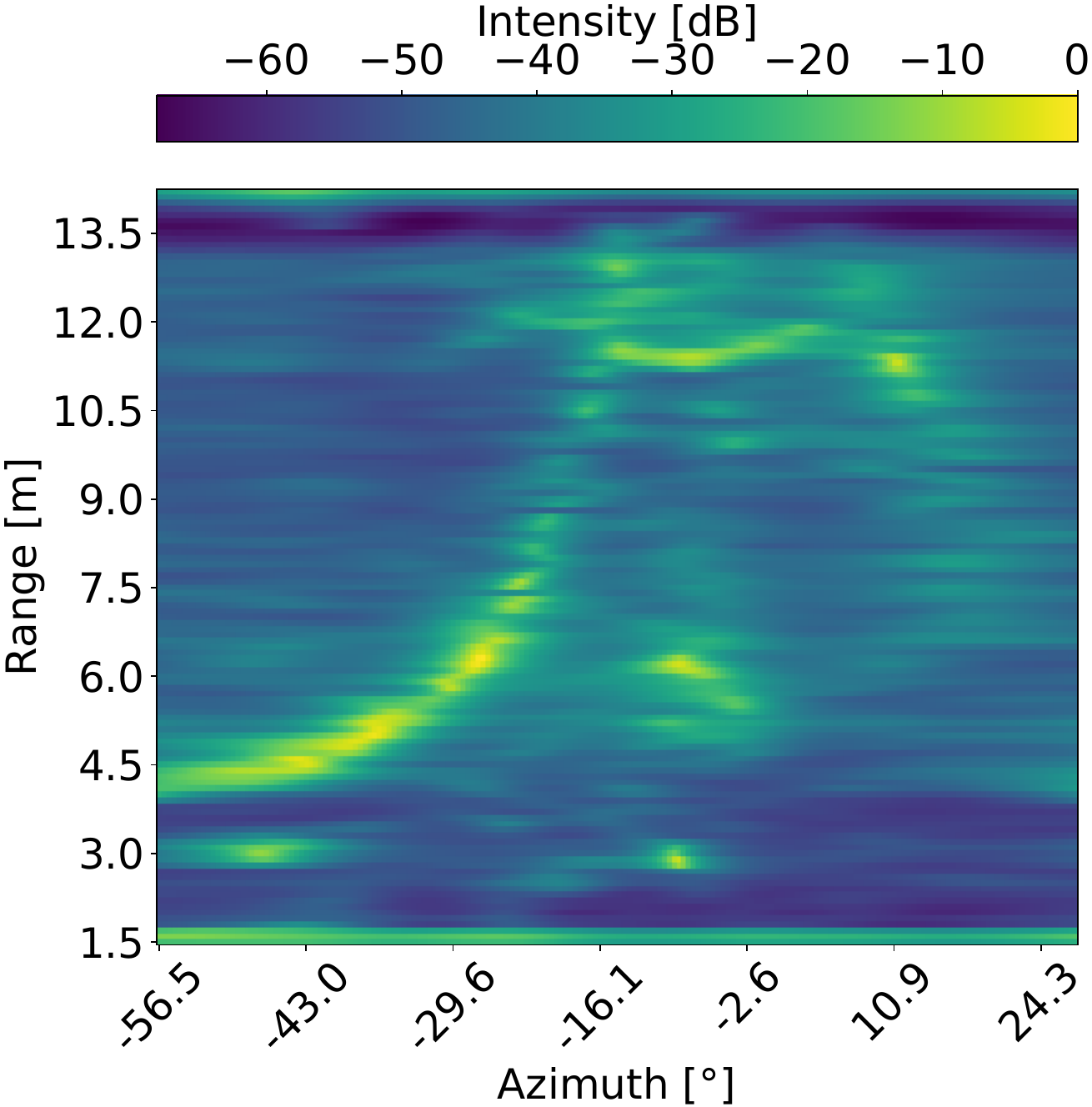}
                \caption{Range-Azimuth Map.}
                \label{fig:ra_map}
        \end{subfigure}
        \begin{subfigure}[b]{0.329\textwidth}
                \includegraphics[width=\textwidth]{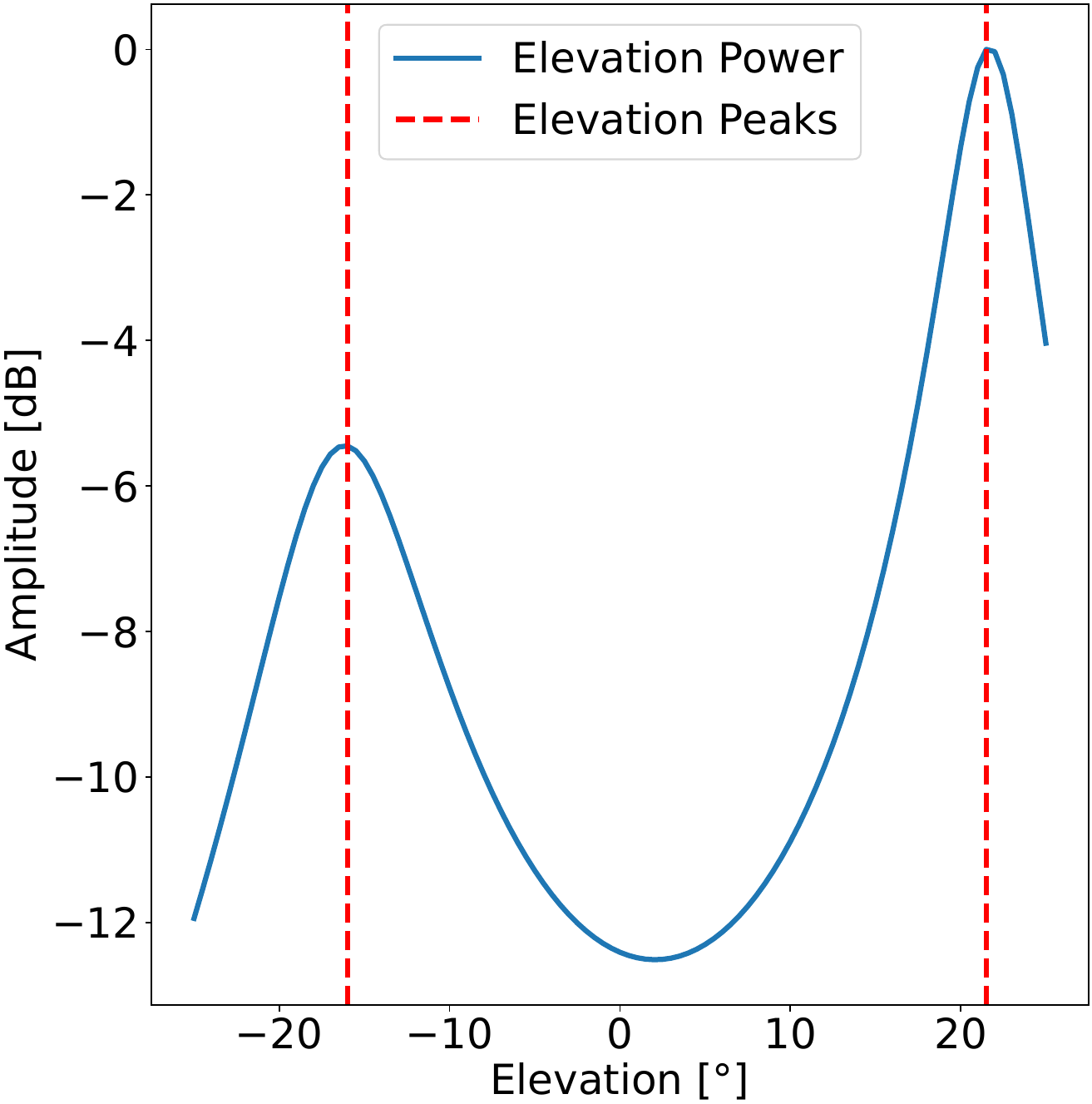}
                \caption{Elevation Spectrum Power.}
                \label{fig:elevation_capon}
        \end{subfigure}
        \begin{subfigure}[b]{0.329\textwidth}
                \includegraphics[width=\textwidth]{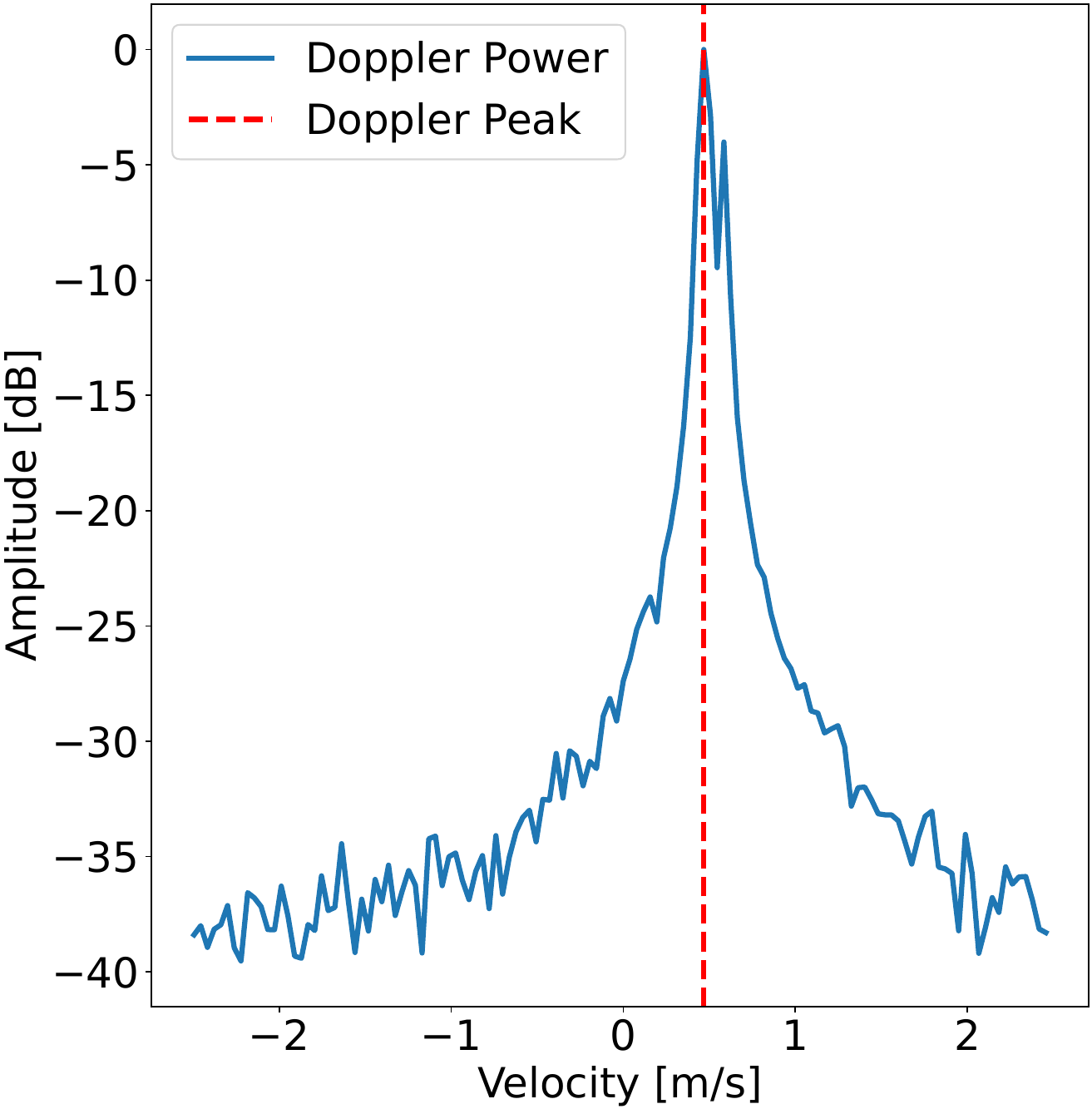}
                \caption{Doppler Spectrum Power.}
                \label{fig:doppler_estimation}
        \end{subfigure}
        \caption{DoA estimation and Doppler estimation.}
        \label{fig:radar_processing}
\end{figure*}

\subsection{DoA Beamforming}
\label{subsec:doa}

After the range-FFT \cite{iovescuFundamentalsMillimeterWave2017},
we estimate the DoA of the potential targets in the spatial domain
using beamforming techniques.
We propose to use a minimum variance distortionless response
(MVDR or Capon) beamforming algorithm \cite{slesickaCaponlikeMethodDirection2023}
as the baseline, and further extend it to 3D
DoA estimation problem.

\subsubsection{Azimuth}
A two-step DoA estimation algorithm is presented in this work,
as illustrated in Figure \ref{fig:system_overview}. First,
we estimate
the DoA of potential targets in the azimuth dimension,
It is generally assumed that the DoA of all targets
relative to the radar remains unchanged within a single frame.
The received signal of all virtual antennas within the frame
can be represented as
\begin{equation}
        \label{eq:received_signal}
        \mathbf{X} = \sum_{n=1}^{N}\begin{bmatrix}
                \mathbf{x}_n(1) & \mathbf{x}_n(2) & \ldots & \mathbf{x}_n(K)
        \end{bmatrix}^{\top},
\end{equation}
where \(N\) is the total number of potential targets in the frame,
\(K\) is the total number of time indexes in the frame,
and \(\mathbf{X} \in \mathbb{R} ^{M \times K}\) is the received
signal matrix.

It can be seen from Eq. \ref{eq:received_signal} that the received
signals from all targets are superimposed. To separate the signal
from different targets, we traverse simultaneously in both
azimuth and range dimensions to calculate the power spectrum
of each azimuth bin and range bin.
Define the self-covariance matrix of the received signal
at the \(j\)-th range bin as
\begin{equation}
        \mathbf{R}_{jj}=\mathbf{X}_j \mathbf{X}^{H}_j/K,
        \label{eq:covariance_matrix}
\end{equation}
where ${\cdot}^{H}$ operation is the conjugate transpose,
and \(\mathbf{X}_j\) is the 2D slice of radar data cube at the \(j\)-th range bin.
Then, the Capon spectrum power $P(\mu)$ of an azimuth bin
\(\mu\) in one range bin can be calculated as
\begin{equation}
        P(\mu) = \frac{1}{\mathbf{a}^H(\mu,\nu) \mathbf{R}_{jj}^{-1} \mathbf{a}(\mu,\nu)}, \quad \text{for } \nu = 0.
\end{equation}
Note that only virtual antennas with zero elevation angle are used
in this step.
We repeat the above process for all azimuth bins and range bins,
which generates a 2D Capon spectrum power map,
also known as a range-azimuth map.
As the range-azimuth map shown in Figure \ref{fig:ra_map}, the pixel color
represents the power of the corresponding azimuth and range.
A brighter pixel indicates a higher possibility of
one or more strong reflection points
in the corresponding azimuth and range. To extract these
potential targets, we apply OS-CFAR detection
\cite{rohlingOrderedStatisticCFAR2011}
on the range-azimuth map
and obtain a
set of azimuth candidates
\(\Theta = \{(\mu_1), (\mu_2), \ldots, (\mu_N)\}\)
for all potential targets.

\subsubsection{Elevation}
Given the set of potential targets \(\Theta\), we estimate the
elevation of each target in the second step using all virtual
antennas.
For the candidate target \((\mu_n)\), the self-covariance
matrix \(\mathbf{R}_{jj}^*\)
is recalculated using Eq. \ref{eq:covariance_matrix},
which incorporates additional virtual antennas with non-zero
elevation angle.
Then, we sweep all possible elevation angles \(\nu\) and calculate
the corresponding Capon spectrum power \(P(\mu_n, \nu)\).
The Capon spectrum power of the \(l\)-th elevation bin \(\nu_l\)
can be calculated as
\begin{equation}
        \left\{
        \begin{aligned}
                P(\mu_n, \nu_l)          & = \frac{1}{\mathbf{a}^H(\mu_n,\nu_l) \mathbf{R}_{jj}^{*-1} \mathbf{a}(\mu_n,\nu_l)} \\
                \mathbf{w}(\mu_n, \nu_l) & = \frac{\mathbf{R}_{jj}^{*-1} \mathbf{a}(\mu_n,\nu_l)}{\mathbf{a}^H(\mu_n,\nu_l)
                        \mathbf{R}_{jj}^{*-1} \mathbf{a}(\mu_n,\nu_l)}
        \end{aligned}
        \right.,
\end{equation}
where \(\mathbf{w}(\mu_n, \nu_l)\) is the Capon beamforming weight vector.

An example of the Capon spectrum power in the elevation dimension
is shown in Figure \ref{fig:elevation_capon}.
The 2D curve in the figure represents the beamforming results
over all possible elevation angles at a given range and azimuth angle.
The peak of the curve indicates the most probable
elevation angle of the target.
Note that there may be multiple peaks in the curve, which
is caused by the multiple targets at the same range and azimuth.
So we extract all the local maximum peaks in the curve as the
elevation angles of the potential targets.
So far, we have obtained the DoA of all potential targets
\( \Theta = \{(\mu_1, \nu_1), (\mu_2, \nu_2),
\ldots, (\mu_N, \nu_N)\} \).

        {\renewcommand{\arraystretch}{1.09}
                \begin{table}[t]
                        \centering
                        \begin{tabular}{cc}
                                \hline
                                \textbf{Parameter}               & \textbf{Value}   \\ \hline
                                Max Azimuth Angular              & ${\pm}$70\degree \\ \hline
                                Max Elevation Angular            & ${\pm}$25\degree \\ \hline
                                Azimuth Bin Number               & 188              \\ \hline
                                Elevation Bin Number             & 101              \\ \hline
                                Azimuth Virtual Antenna number   & 8                \\ \hline
                                Elevation Virtual Antenna number & 4                \\ \hline
                        \end{tabular}
                        \caption{Parameter Specifications}
                        \label{tab:specifications}
                \end{table}
        }
\subsection{Doppler Estimation}
\label{subsec:Doppler}
After the DoA estimation, we estimate the Doppler of all targets.
Given the DoA of the \(n\)-th target \((\mu_n, \nu_n)\),
we need to suppress the interference of other targets.
This can be achieved by applying the Capon beamforming on the
\(\mathbf{X}^{*}_j\) using the Capon beamforming weight vector
\(\mathbf{w}(\mu_n, \nu_n)\).
The beamforming results of the \(n\)-th target is
\begin{equation}
        \mathbf{x}_n = \mathbf{w}^H(\mu_n, \nu_n) \mathbf{X}^{*}_j
\end{equation}
where \(\mathbf{x}_n \in \mathbb{R}^{1 \times K}\) only contains
the Doppler information of the \(n\)-th target.
Then  1D FFT is applied on \(\mathbf{x}_n\) to estimate the
Doppler of the \(n\)-th target.
The Doppler spectrum power of the
\(n\)-th target is given by
\begin{equation}
        P_n = |\text{FFT}(x_n)|.
\end{equation}
We can now find
the maximum of the Doppler spectrum power by
$V_{\text{Doppler}} = \arg\max_{V} P_n(V)$
as  the target's estimated  radial velocity \(V_{\text{Doppler}}\),
as the example shown in Figure \ref{fig:doppler_estimation}.

Given these detected 3D target points and their associated Doppler velocities, we propose to use them, as more accurate replacements for those generated from the standard FFT-based pipeline, with radar odometry/SLAM algorithms.

        {\renewcommand{\arraystretch}{1.1}
                \begin{table}[t]
                        \centering
                        \caption{Chamfer Distance for all point clouds [m]}
                        \begin{tabular}{llcccc}
                                \toprule
                                \multirow{2}{*}{\textbf{Res.}} & \multirow{2}{*}{\textbf{Mode}} & \multicolumn{4}{c}{\textbf{Sequence}}                                                          \\
                                                               &                                & \textbf{hallways0}                    & \textbf{outdoors0} & \textbf{aspen11} & \textbf{army2} \\ \hline

                                \multirow{2}{*}{0.2}           & Ours                           & \textbf{0.14560}                      &
                                \textbf{0.15477}               & \textbf{0.16951}               & \textbf{0.15622}                                                                               \\
                                                               & FFT                            & 0.16443                               &
                                0.15828                        & 0.17146                        & 0.16803                                                                                        \\ \hline
                                \multirow{2}{*}{0.1}           & Ours                           & \textbf{0.05194}                      &
                                \textbf{0.05834}               & \textbf{0.06198}               & \textbf{0.05277}                                                                               \\
                                                               & FFT                            & 0.06566                               &
                                0.06308                        & 0.07320                        & 0.06703                                                                                        \\ \hline
                                \multirow{2}{*}{0.05}          & Ours                           & \textbf{0.01945}                      &
                                \textbf{0.02162}               & \textbf{0.02333}               & \textbf{0.01831}                                                                               \\
                                                               & FFT                            & 0.02477                               &
                                0.02418                        & 0.02733                        & 0.02475                                                                                        \\ \hline
                                \multirow{2}{*}{0.02}          & Ours                           & \textbf{0.00322}                      &
                                \textbf{0.00381}               & \textbf{0.00646}               & \textbf{0.00346}                                                                               \\
                                                               & FFT                            & 0.00421                               &
                                0.00440                        & 0.00699                        & 0.00453                                                                                        \\ \hline
                                \multirow{2}{*}{0.01}          & Ours                           & \textbf{0.00065}                      &
                                \textbf{0.00072}               & \textbf{0.00154}               & \textbf{0.00072}                                                                               \\
                                                               & FFT                            & 0.00084                               &
                                0.00085                        & 0.00168                        & 0.00090                                                                                        \\
                                \bottomrule
                        \end{tabular}
                        \label{tab:fft_capon_chamfer_distance_separated}
                \end{table}
        }

\section{EXPERIMENTS}

 {\renewcommand{\arraystretch}{1.12}
\begin{table*}[t]
        \centering
        \caption{Drift with per distance and APE (RMSE) for each sequence. Red highlights the best result on each sequence. }
        \resizebox{\textwidth}{!}{%
                \begin{tabular}{ccccccccc}
                        \hline
                        \multirow{2}{*}{\textbf{Sequence}}                                            & \multirow{2}{*}{\textbf{RIO}} & \multirow{2}{*}{\textbf{Point cloud}} & \multicolumn{4}{c}{\textbf{Drift  with per distance}} & \multicolumn{2}{c}{\textbf{APE (RMSE)}}                                                                                                                                                          \\
                                                                                                      &                               &                                       & \textbf{XYZ}{[}m{]}{[}\%{]}                           & \textbf{XY}{[}m{]}{[}\%{]}              & \textbf{Z}{[}m{]}{[}\%{]}             & \textbf{Yaw}{[}\degree{]}{[}{\degree}/m{]} & \textbf{Trans.} {[}m{]}         & \textbf{Yaw.}{[}\degree{]}      \\ \hline
                        \multirow{4}{*}{\begin{tabular}[c]{@{}c@{}}outdoors0\\ 115.4m\end{tabular}}   & \multirow{2}{*}{Graph}        & Ours                                  & 5.658     [4.9]                                       & \textcolor{red}{\textbf{0.473 [0.4]}}   & 5.638 [4.8]                           & 1.140   [0.010]                            & 1.807                           & 0.158                           \\
                                                                                                      &                               & FFT                                   & 6.087     [5.2]                                       & 1.327 [1.1]                             & 5.940 [5.1]                           & 0.984   [0.009]                            & 2.377                           & \textcolor{red}{\textbf{0.093}} \\ \cdashline{2-9}
                                                                                                      & \multirow{2}{*}{EKF}          & Ours                                  & \textcolor{red}{\textbf{1.513     [1.3]}}             & 0.542 [0.4]                             & \textcolor{red}{\textbf{1.413 [1.2]}} & \textcolor{red}{\textbf{0.816   [0.007]}}  & \textcolor{red}{\textbf{0.678}} & 0.172                           \\
                                                                                                      &                               & FFT                                   & 13.426    [11.6]                                      & 3.210 [2.7]                             & 13.037 [11.2]                         & 0.892   [0.008]                            & 4.314                           & 0.172                           \\ \hline
                        \multirow{4}{*}{\begin{tabular}[c]{@{}c@{}}aspen11\\ 78.9m\end{tabular}}      & \multirow{2}{*}{Graph}        & Ours                                  & 7.791     [9.8]                                       & \textcolor{red}{\textbf{0.161 [0.2]}}   & 7.789 [9.8]                           & 0.482   [0.006]                            & 2.488                           & 0.144                           \\
                                                                                                      &                               & FFT                                   & \textcolor{red}{\textbf{5.482     [6.9]}}             & 0.883 [1.1]                             & \textcolor{red}{\textbf{5.410 [6.8]}} & 0.493   [0.006]                            & \textcolor{red}{\textbf{1.704}} & \textcolor{red}{\textbf{0.101}} \\ \cdashline{2-9}
                                                                                                      & \multirow{2}{*}{EKF}          & Ours                                  & 6.821     [8.6]                                       & 0.301 [0.3]                             & 6.815 [8.6]                           & \textcolor{red}{\textbf{0.318   [0.004]}}  & 2.264                           & 0.149                           \\
                                                                                                      &                               & FFT                                   & 10.768    [13.6]                                      & 0.992 [1.2]                             & 10.722 [13.5]                         & 0.580   [0.007]                            & 3.314                           & 0.133                           \\ \hline
                        \multirow{4}{*}{\begin{tabular}[c]{@{}c@{}}{army2}\\ 129.2m\end{tabular}}     & \multirow{2}{*}{Graph}        & Ours                                  & \textcolor{red}{\textbf{2.959     [2.2]}}             & 2.368 [1.8]                             & \textcolor{red}{\textbf{1.775 [1.3]}} & \textcolor{red}{\textbf{0.400   [0.003]}}  & \textcolor{red}{\textbf{1.281}} & 0.149                           \\
                                                                                                      &                               & FFT                                   & 18.889    [14.5]                                      & 1.041 [0.8]                             & 18.861 [14.5]                         & 0.825   [0.006]                            & 6.114                           & \textcolor{red}{\textbf{0.122}} \\ \cdashline{2-9}
                                                                                                      & \multirow{2}{*}{EKF}          & Ours                                  & 5.724     [4.4]                                       & 1.483 [1.1]                             & 5.528 [4.2]                           & 1.384   [0.011]                            & 2.051                           & 0.209                           \\
                                                                                                      &                               & FFT                                   & 23.159    [17.8]                                      & \textcolor{red}{\textbf{0.796 [0.6]}}   & 23.145 [17.8]                         & 0.602   [0.005]                            & 7.359                           & 0.157                           \\ \hline
                        \multirow{4}{*}{\begin{tabular}[c]{@{}c@{}}{hallways0}\\ 111.4m\end{tabular}} & \multirow{2}{*}{Graph}        & Ours                                  & 1.980     [1.7]                                       & \textcolor{red}{\textbf{0.650 [0.5]}}   & 1.870 [1.6]                           & 0.402   [0.004]                            & 1.450                           & 0.167                           \\
                                                                                                      &                               & FFT                                   & 15.488    [13.8]                                      & 2.103 [1.8]                             & 15.344 [13.7]                         & \textcolor{red}{\textbf{0.398   [0.004]}}  & 4.667                           & \textcolor{red}{\textbf{0.159}} \\ \cdashline{2-9}
                                                                                                      & \multirow{2}{*}{EKF}          & Ours                                  & \textcolor{red}{\textbf{1.360     [1.2]}}             & 0.883 [0.7]                             & \textcolor{red}{\textbf{1.035 [0.9]}} & 1.561   [0.014]                            & \textcolor{red}{\textbf{1.337}} & 0.171                           \\
                                                                                                      &                               & FFT                                   & 15.879    [14.2]                                      & 1.423 [1.2]                             & 15.815 [14.1]                         & 1.925   [0.017]                            & 5.007                           & 0.162                           \\ \hline
                \end{tabular}%
        }
        \label{tab:methods_comparison_rel}
\end{table*}
}

\subsection{Experimental Setup}
To evaluate the performance of the proposed framework, we conduct a series of
experiments on the public ColoRadar dataset \cite{kramerColoRadarDirect3D2022}.
The raw radar data in the dataset is used %
to evaluate the performance.
The raw radar data is collected by a single-chip SoC
radar (Texas Instruments AWR1843BOOST-EVM paired with a DCA1000EVM).
The proposed pipeline processes the raw radar data to generate
an accurate radar point cloud (called our point cloud).
The radar point clouds, provided by the ColoRadar dataset, are generated by the
traditional FFT pipeline (called FFT point cloud).
These two different pipelines are compared in four different sequences:
ec\_hallways\_run0, outdoors\_run0, aspen\_run11, and edgar\_army\_run2,
which contains indoor and outdoor scenes, structural and unstructured scenes.
Key parameters of the proposed pipeline are shown in Table \ref{tab:specifications}.
We do not compare learning-based methods in this work, as they can not provide
the Doppler information of the radar point cloud, which is essential
for radar-inertial odometry.

\begin{figure}[t]
        \centering
        \vspace{-0.1cm}
        \begin{subfigure}[b]{0.485\textwidth}
                \includegraphics[width=\textwidth]{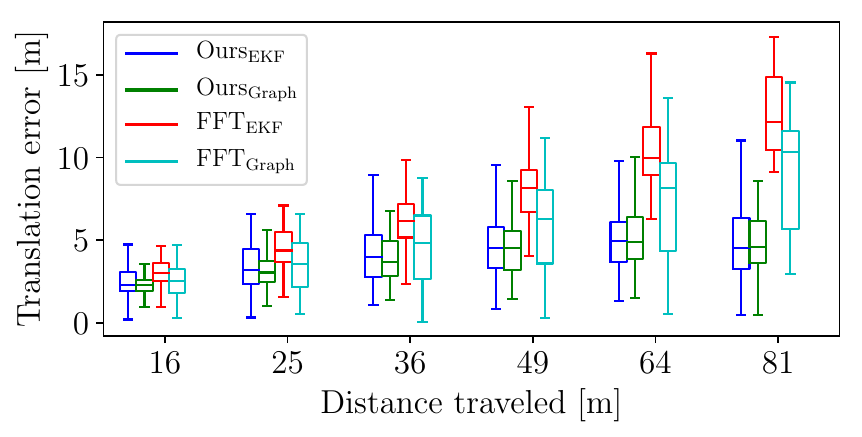}
                \caption{Translation}
        \end{subfigure}
        \hfill

        \vspace{-0.5cm}
        \begin{subfigure}[b]{0.485\textwidth}
                \includegraphics[width=\textwidth]{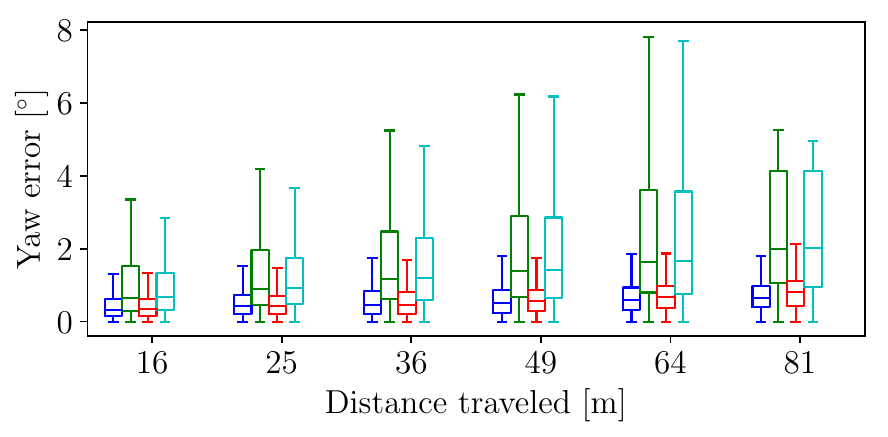}
                \caption{Rotation}
        \end{subfigure}
        \caption{Boxplot of the RPE in all sequences.}
        \label{fig:rpe_boxplot}
\end{figure}

\subsection{Point Precision}

In order to analyze the point precision, we first generate groundtruth
environmental point clouds by projecting all LiDAR point clouds into the global
frame.
Then we project the FFT and our radar point clouds using the provided groundtruth poses
and the extrinsic parameters between radar and Vicon.
The Chamfer
Distance \cite{jiaoFusionPortableMultiSensorCampusScene2022} is calculated
using the estimated point cloud and the groundtruth point cloud.
The result is shown in Table \ref{tab:fft_capon_chamfer_distance_separated}.
{The parameter Res. represents the resolution level of the point clouds, affecting the density of detected points before Chamfer Distance computation.}

In all scenes and resolutions, our point clouds have a smaller Chamfer Distance value
than the FFT point clouds.
This metric decreased by 1.14\% to 26.02\%, with an average decrease of 14.70\%.
A smaller value of Chamfer Distance
indicates that the estimated point cloud has a smaller error
to the groundtruth point cloud. Thus, our point cloud is closer to
the groundtruth point cloud in all scenes and resolutions.
This result demonstrates that the proposed pipeline can
generate more accurate point clouds than the traditional FFT pipeline.

\subsection{Odometry Accuracy}

\begin{figure*}[htbp]
        \centering
        \begin{subfigure}[b]{0.245\textwidth}
                \includegraphics[width=\textwidth]{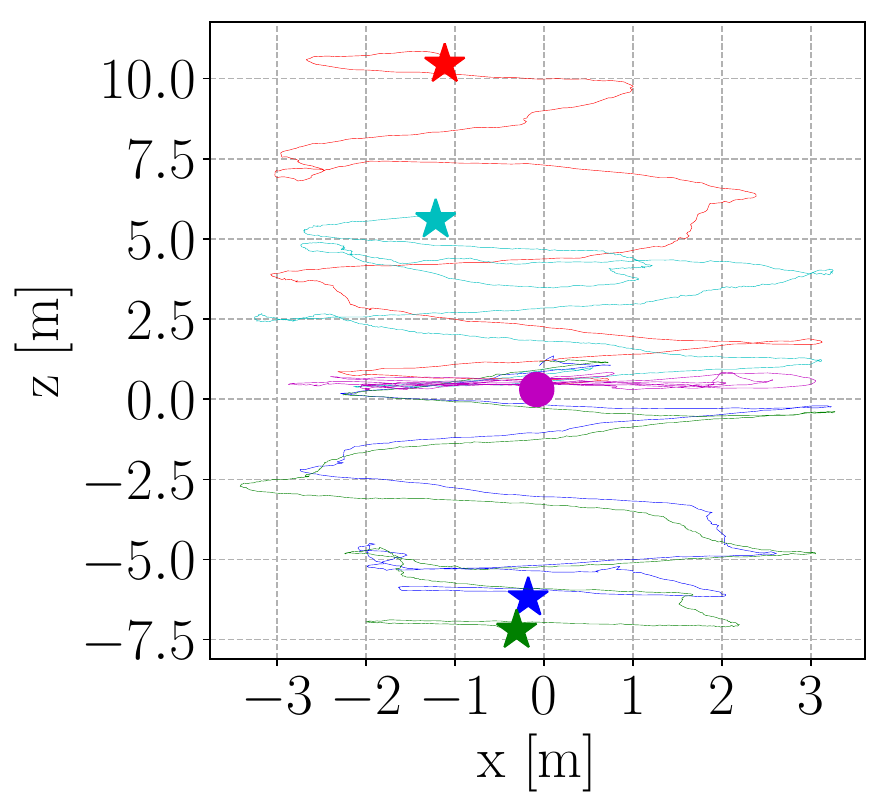}
                \caption{aspen11}
        \end{subfigure}
        \begin{subfigure}[b]{0.245\textwidth}
                \includegraphics[width=\textwidth]{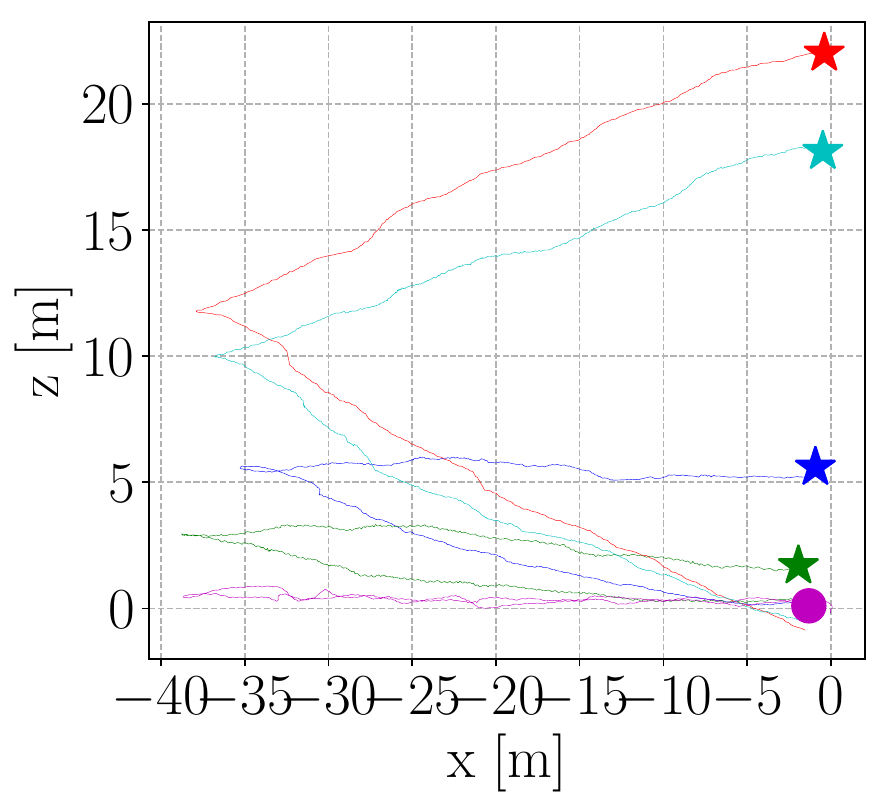}
                \caption{army2}
        \end{subfigure}
        \begin{subfigure}[b]{0.245\textwidth}
                \includegraphics[width=\textwidth]{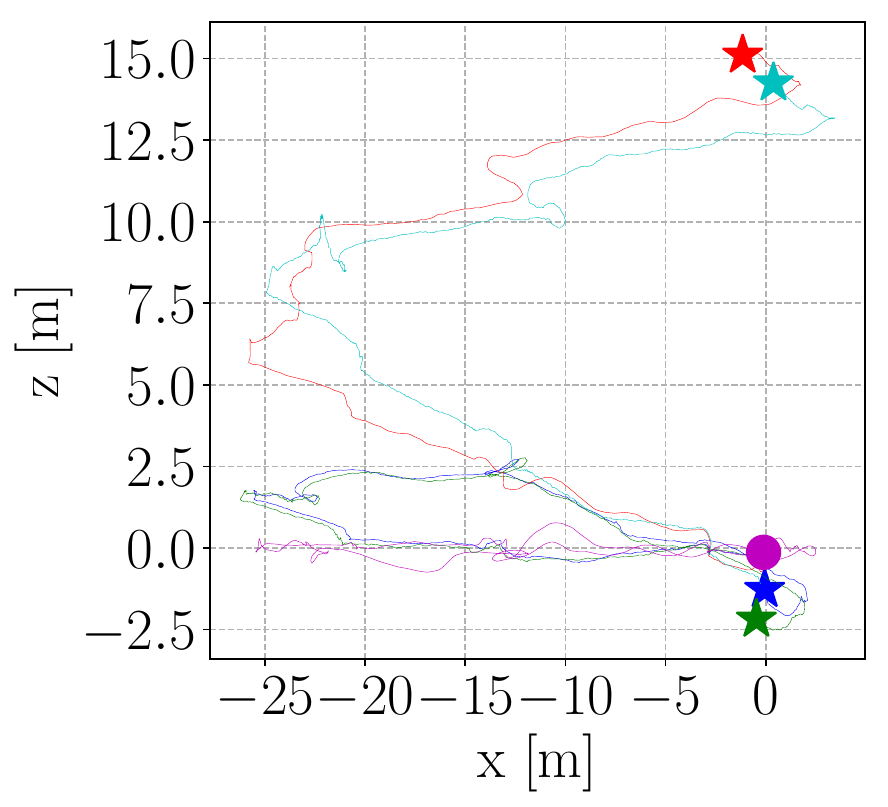}
                \caption{hallways0}
        \end{subfigure}
        \begin{subfigure}[b]{0.245\textwidth}
                \includegraphics[width=\textwidth]{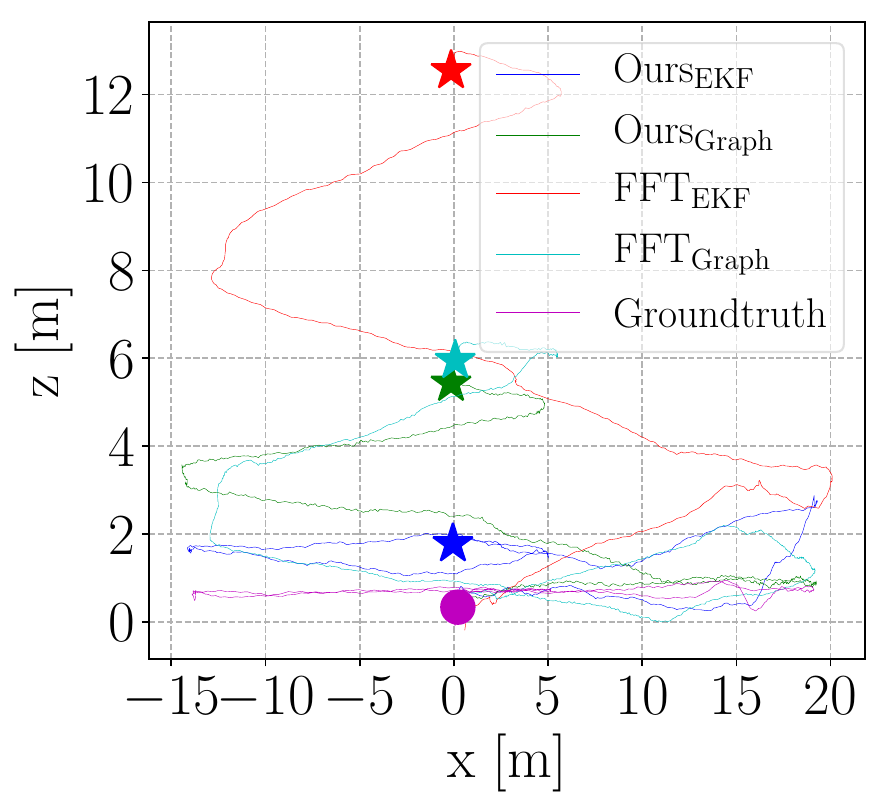}
                \caption{outdoors0}
        \end{subfigure}
        \caption{Side view of all trajectories in the XZ-plane of all sequences.
                The dots represent the starting positions, while the pentagrams represent the end positions.
                {The displacement between them indicates the degree of drift in each trajectory.}
        }
        \label{fig:side_trajectory_all}
\end{figure*}

\begin{figure*}[htpb]
        \centering
        \begin{subfigure}[b]{0.245\textwidth}
                \includegraphics[width=\textwidth]{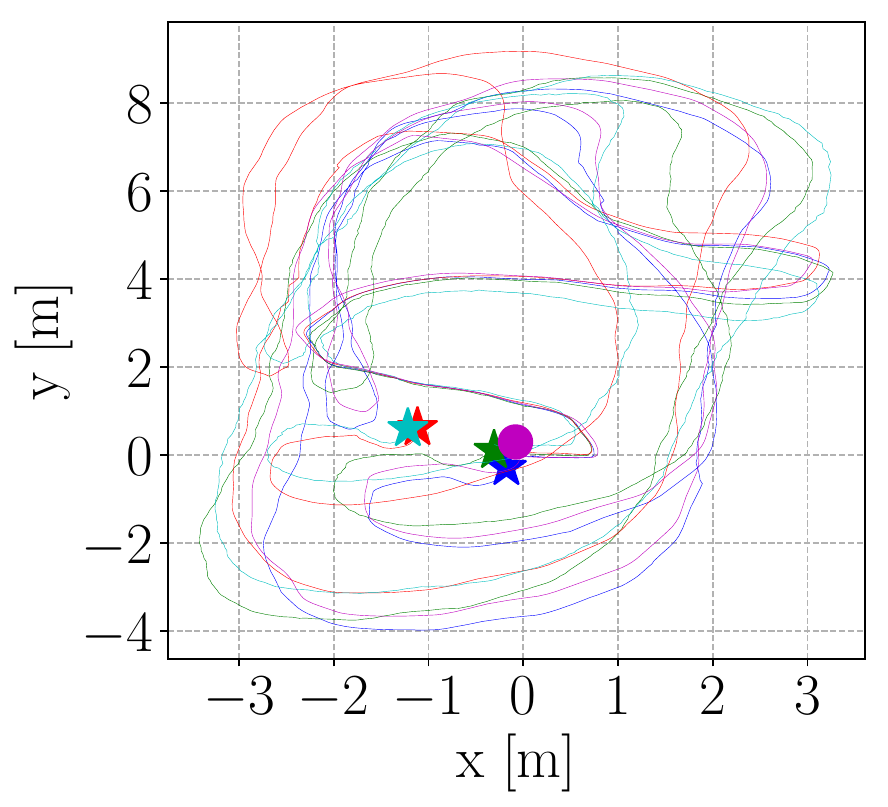}
                \caption{aspen11}
        \end{subfigure}
        \hfill
        \begin{subfigure}[b]{0.245\textwidth}
                \includegraphics[width=\textwidth]{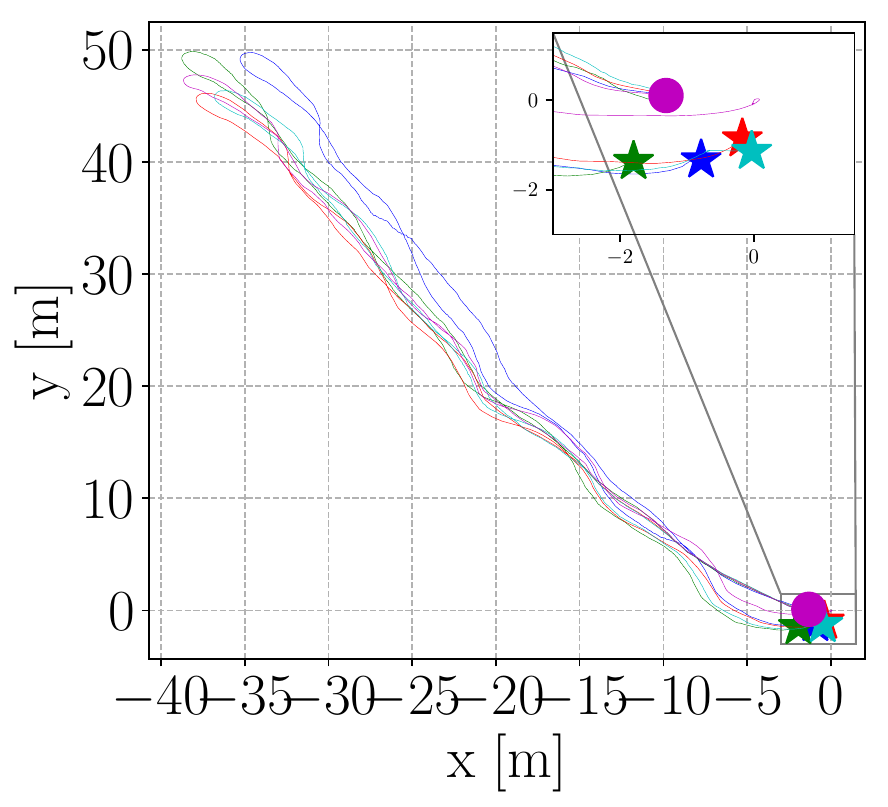}
                \caption{army2}
        \end{subfigure}
        \hfill
        \begin{subfigure}[b]{0.245\textwidth}
                \includegraphics[width=\textwidth]{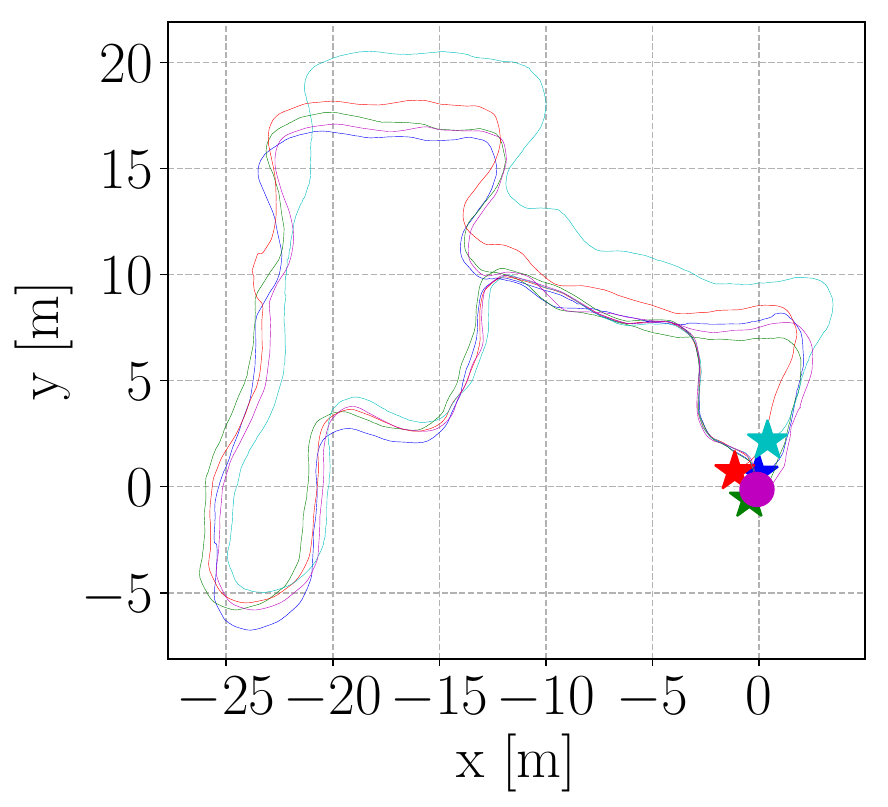}
                \caption{hallways0}
        \end{subfigure}
        \hfill
        \begin{subfigure}[b]{0.245\textwidth}
                \includegraphics[width=\textwidth]{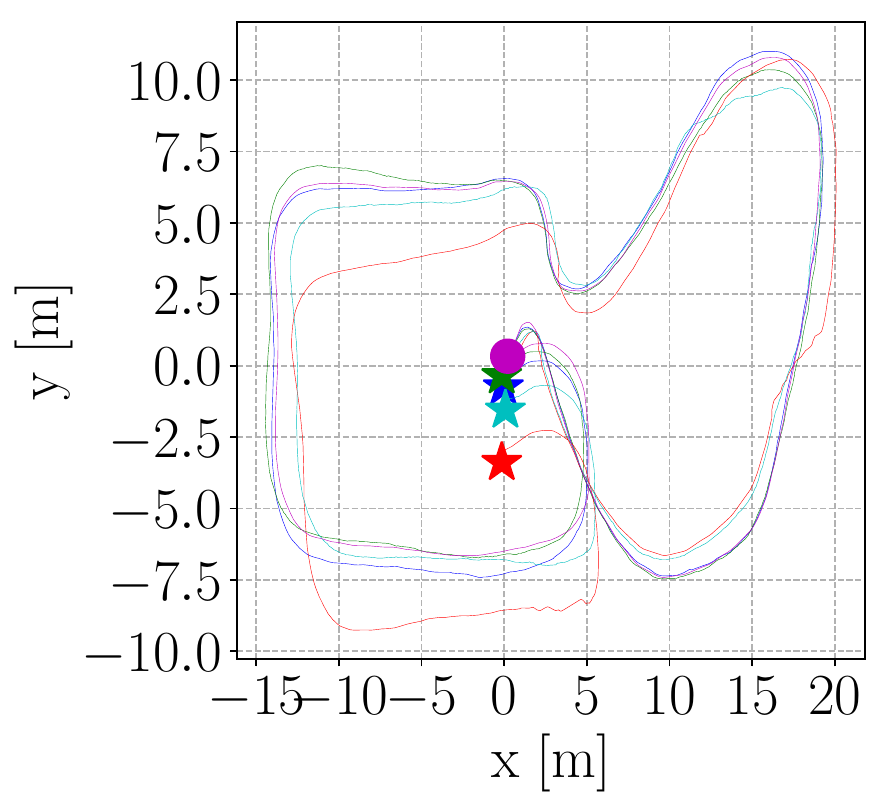}
                \caption{outdoors0}
        \end{subfigure}
        \caption{Top view of all trajectories in the XY-plane of all sequences.
                The colors of the trajectories, as well as dots and pentagrams,
                are consistent with those used in the previous figures.
        }

        \label{fig:top_trajectory_all}
\end{figure*}

To further evaluate the influence of the proposed pipeline on a full
radar-inertial odometry system,
we select two different radar-inertial odometry systems as the baseline,
and compare their localization accuracy incorporating different radar
signal processing pipelines.
The first system is a filter-based radar-inertial odometry
\cite{doerEKFBasedApproach2020}. It calculates the ego-velocity of
a single frame using Doppler information of all radar
point clouds within this frame,
and then fuses the ego-velocity with the IMU data
in an Extended Kalman Filter to estimate the pose.
The second system is the graph-based radar-inertial odometry
\cite{girodRobustBaroradarinertialOdometry2024}.
Rather than calculating the ego-velocity of a single frame,
it directly uses the Doppler information
of single radar point as the edge in the graph, and then
optimizes the pose and velocity of all frames in a fixed lag
sliding window. These two radar-inertial odometry systems
stand for the two main fusion strategies in radar-inertial odometry.

To make a fair comparison, we only adjust the radar-related parameters
in these two systems, and keep all other parameters unchanged.
The barometer fusion strategy is not activated in both systems,
due to the lack of barometer data in the ColoRadar dataset.
We evaluate the odometry accuracy using the Absolute Pose Error (APE)
and Relative Pose Error (RPE) metrics \cite{zhangTutorialQuantitativeTrajectory2018}.
The statistics are shown in Table \ref{tab:methods_comparison_rel}.
The boxplot of the RPE summarized in
all sequences are shown in Figure \ref{fig:rpe_boxplot}.

It can be seen that the proposed pipeline
can significantly improve the localization performance of different
radar-inertial odometry systems in different scenes.
The translation error decreases by 1.62\% to 22.07\%,
with an average decrease of 13.68\%. The rotation error does not have
a significant change.
This may be due to the fact that these two radar-inertial odometry
systems mainly use the Doppler information, rather than the structure information.
The Doppler information mostly affects
the translation estimation and has less impact on the rotation estimation.
Thus, even though the proposed pipeline can generate more accurate point clouds,
Doppler-only update strategy in these two systems may not fully utilize
the improved point clouds, which needs further investigation.

To further analyze the odometry accuracy, we visualize the estimated
trajectories in the XZ-plane and XY-plane, as shown in
Figure \ref{fig:side_trajectory_all} and Figure \ref{fig:top_trajectory_all}.
In the both XY-plane and XZ-plane, the estimated
trajectories based on the proposed pipeline
are closer to the groundtruth.
The first reason is that the proposed pipeline mainly focuses on extracting the stable and
strong reflection points in the radar signal
while ignoring the noisy and weak reflection points.
This characteristic can decrease the possibility of generating outliers or noise points
in the radar point cloud, which in turn can improve the odometry accuracy.
The second reason is traditional FFT pipeline needs to compensate for
the Doppler estimation before the DoA estimation,
which may introduce additional errors in the DoA estimation.
Furthermore, the single-chip radar used in the ColoRadar dataset
only has two virtual elevation antennas, which further exacerbates
the traditional pipeline elevation errors.
This could explain why the proposed pipeline can significantly decrease the
Z-axis drift in the estimated trajectory, as shown in Figure \ref{fig:side_trajectory_all}.

\section{Conclusion}
In this paper, we design a novel radar signal processing pipeline.
Taking full advantage of the digital beamforming techniques, the proposed
pipeline can generate more accurate 3D radar
point clouds than traditional FFT-based
pipelines. The proposed pipeline is evaluated on the public ColoRadar dataset,
and the results show that the proposed pipeline can significantly decrease
the error of the radar point clouds, and further improve the accuracy of
radar-inertial odometry systems.
This paper highlights the significance of utilizing raw radar signals
from a single-chip radar to enhance the localization and mapping accuracy in radar-inertial odometry systems.
This approach is akin to designing more advanced feature extraction algorithms for visual odometry/SLAM.
Future research can focus on improving the rotation estimation accuracy.

\cleardoublepage

\bibliographystyle{IEEEtran.bst}
\bibliography{new}

\end{document}